\documentclass[conference]{IEEEtran}

\IEEEoverridecommandlockouts
\usepackage{cite}
\usepackage{amsmath,amssymb,amsfonts}
\usepackage{algorithmic}
\usepackage{graphicx}
\usepackage{textcomp}
\usepackage{xcolor}
\usepackage{booktabs}
\usepackage{multirow}
\usepackage{tcolorbox}
\usepackage[hyphens]{url}
\usepackage{hyperref}
\usepackage[none]{hyphenat}

\def\BibTeX{{\rm B\kern-.05em{\sc i\kern-.025em b}\kern-.08em
    T\kern-.1667em\lower.7ex\hbox{E}\kern-.125emX}}

\begin{document}

\title{MedTrust-RAG: Evidence Verification and Trust Alignment for Biomedical Question Answering
}

\author{\IEEEauthorblockN{Yingpeng Ning\textsuperscript{1}, Yuanyuan Sun\textsuperscript{1}, Ling Luo\textsuperscript{1,*}, Yanhua Wang\textsuperscript{2}, Yuchen Pan\textsuperscript{1}, Hongfei Lin\textsuperscript{1}
\IEEEauthorblockA{\textsuperscript{1}College of Computer Science and Technology, Dalian University of Technology, Dalian, China \\
\textsuperscript{2}Air Force Communications NCO Academy, Dalian, China
\\
\textsuperscript{*}To whom correspondence should be addressed: lingluo@dlut.edu.cn
}
}}

\maketitle

\begin{abstract}
Biomedical question answering (QA) requires accurate interpretation of complex medical knowledge. Large language models (LLMs) have shown promising capabilities in this domain, with retrieval-augmented generation (RAG) systems enhancing performance by incorporating external medical literature. However, RAG-based approaches in biomedical QA suffer from hallucinations due to post-retrieval noise and insufficient verification of retrieved evidence, undermining response reliability. We propose \textbf{MedTrust-Guided Iterative RAG}, a framework designed to enhance factual consistency and mitigate hallucinations in medical QA. Our method introduces three key innovations. First, it enforces citation-aware reasoning by requiring all generated content to be explicitly grounded in retrieved medical documents, with structured Negative Knowledge Assertions used when evidence is insufficient. Second, it employs an iterative retrieval-verification process, where a verification agent assesses evidence adequacy and refines queries through Medical Gap Analysis until reliable information is obtained. Third, it integrates the \textbf{MedTrust-Align Module (MTAM)} that combines verified positive examples with hallucination-aware negative samples, leveraging Direct Preference Optimization to reinforce citation-grounded reasoning while penalizing hallucination-prone response patterns. Experiments on MedMCQA, MedQA, and MMLU-Med demonstrate that our approach consistently outperforms competitive baselines across multiple model architectures, achieving the best average accuracy with gains of 2.7\% for LLaMA3.1-8B-Instruct and 2.4\% for Qwen3-8B.
\end{abstract}


\section{Introduction}

Biomedical question answering (QA) is a critical task that requires accurate understanding of complex medical knowledge, effective synthesis of evidence from diverse sources, and precise reasoning over specialized clinical concepts to support healthcare decision-making. Large language models (LLMs) have demonstrated exceptional capabilities in this domain, excelling in medical question answering, clinical decision support, and biomedical text analysis~\cite{anil2023palm}. However, their deployment in critical medical environments remains constrained due to requirements for high factual accuracy and specialized domain reasoning. A primary concern is the phenomenon of hallucination, where models generate plausible-sounding but factually incorrect information~\cite{zhang2025siren}. In clinical contexts, such inaccuracies can result in unsafe recommendations, diminish clinician trust, and hinder widespread adoption~\cite{hersh2024search}. Furthermore, the static nature of biomedical knowledge embedded in LLMs makes it challenging and expensive to incorporate current medical advances~\cite{kasai2023realtime}. To address these limitations, retrieval-augmented generation (RAG)~\cite{lewis2020retrieval} offers a promising approach by dynamically retrieving relevant biomedical literature for each query, enabling LLMs to access current and accurate information while enhancing factual reliability and reducing hallucination rates~\cite{xiong2024benchmarking}.

\begin{figure}[t]
\centering
\includegraphics[scale=0.6]{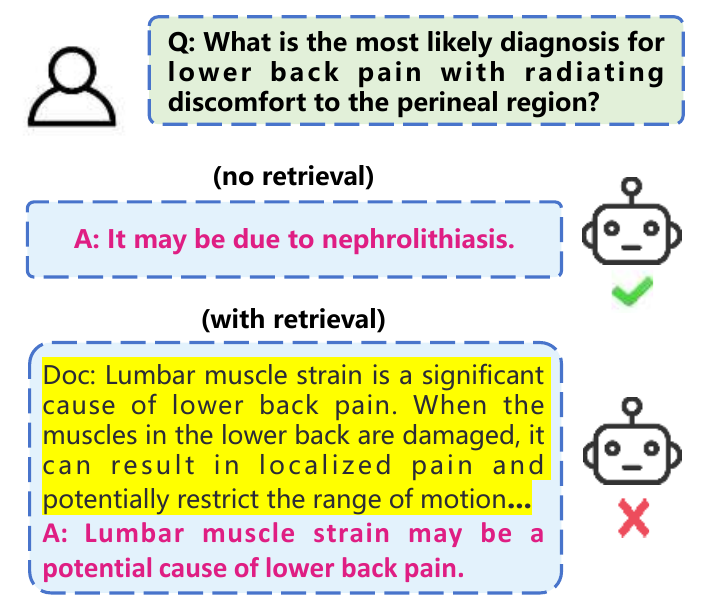}
\caption{An example illustrating post-retrieval noise in biomedical RAG. The model answers the question correctly without retrieval, but produces an incorrect response when using retrieved documents, highlighting the potential risk of introducing irrelevant or misleading information.}
\label{f-1}
\end{figure}

However, applying RAG in the biomedical domain presents several unique challenges. First, irrelevant or misleading retrieved content can significantly misdirect LLMs~\cite{han2023medalpaca}. Biomedical literature often involves complex terminology and nuanced clinical concepts. As a result, retrieval methods based only on semantic similarity may retrieve documents that seem relevant on the surface but fail to provide clinically meaningful information~\cite{dai2024neural}. When such content is included in the input, it may lead the model to focus on less relevant information and undermine the accuracy of the response. Fig.~\ref{f-1} illustrates this issue, where the introduction of retrieved documents leads the model to change an originally correct answer to an incorrect one due to misleading information.

Second, retrieved documents often fail to match the knowledge stored in the model’s parameters. When the generated answer conflicts with the retrieved content, hallucinations become more likely. Studies have shown that large language models tend to rely heavily on internal knowledge and may ignore external evidence, even when that evidence is accurate~\cite{sun2024redeep}. In addition, most RAG systems lack effective mechanisms to evaluate the relevance and reliability of retrieved information. As a result, irrelevant or misleading passages are often used without proper validation~\cite{pham2024towards}.

To address these challenges, we propose\textbf{ MedTrust-Guided Iterative RAG} (Medical Trust-Guided Iterative Retrieval Augmented Generation), a medical trust-aligned retrieval-augmented framework designed to mitigate irrelevant knowledge introduction and reduce hallucination in medical question answering. First, we implement citation-aware medical knowledge statements that require all reasoning content to be explicitly traceable to retrieved medical documents. This mechanism ensures each explanatory statement is substantiated by empirical evidence and accompanied by precise inline citations linking to specific source documents. When retrieved documents lack sufficient evidentiary support for medically reliable responses, our approach executes a principled refusal protocol through structured \textbf{Negative Knowledge Assertions} rather than attempting synthesis from inadequate evidence. Second, we employ an iterative retrieval-verification pipeline with a specialized verification agent that continuously evaluates whether retrieved content provides adequate support for sophisticated medical questions. When evidence gaps are identified, the framework dynamically refines retrieval queries with \textbf{Medical Gap Analysis} until sufficient knowledge is obtained. Third, to improve alignment with medical domain requirements, we introduce a medical trust alignment methodology that combines verified positive samples with hallucination-aware negative sampling. This training strategy addresses four critical hallucination patterns in biomedical contexts through systematic construction of negative samples and multi-model collaboration, utilizing direct preference optimization (DPO)~\cite{rafailov2023direct} to reinforce preference for verified citation-grounded reasoning while penalizing hallucination-prone response patterns.

We evaluate the MedTrust-Guided Iterative RAG framework on three open-domain biomedical QA benchmarks, including MedMCQA~\cite{pal2022medmcqa}, MedQA~\cite{jin2021disease}, and MMLU-Med~\cite{hendrycks2021measuring}. Our method substantially improves the average accuracy of strong LLMs across different architectures, such as LLaMA3.1-8B-Instruct~\cite{touvron2024llama3} and Qwen3-8B~\cite{qwen32024}. Compared to the strongest standard RAG baseline, our framework yields absolute gains of 2.7~\% on LLaMA3.1-8B-Instruct and 2.4~\% on Qwen3-8B. Furthermore, the DPO-trained model consistently outperforms supervised fine-tuning (SFT)~\cite{radford2019language}, demonstrating the effectiveness of medical trust alignment in biomedical question answering.

Our contributions are summarized as follows:

\begin{itemize}
\item \textbf{Iterative Retrieval-Verification Pipeline}. We introduce a dual-agent architecture with specialized verifier and generator agents. The verifier evaluates medical validity of retrieved evidence and generates citation-grounded reasoning or refusal statements, while the generator produces answers exclusively from validated inputs.

\item \textbf{Medical Trust Alignment Framework}. We develop a MedTrust-Align methodology that combines verified positive samples with hallucination-aware negative sampling, utilizing DPO to reinforce citation-grounded reasoning while penalizing hallucination patterns.

\item \textbf{Comprehensive Empirical Validation}. Experiments on MedMCQA, MedQA, and MMLU-Med demonstrate that our framework consistently outperforms strong baselines under both LLaMA3.1-8B and Qwen3-8B. Compared to the strongest RAG-based method, it achieves up to +2.7~\% absolute gains in average accuracy.
\end{itemize}

\section{Methods}
\label{sec:method}

\begin{figure*}[t]
\centering
\includegraphics[scale=0.62]{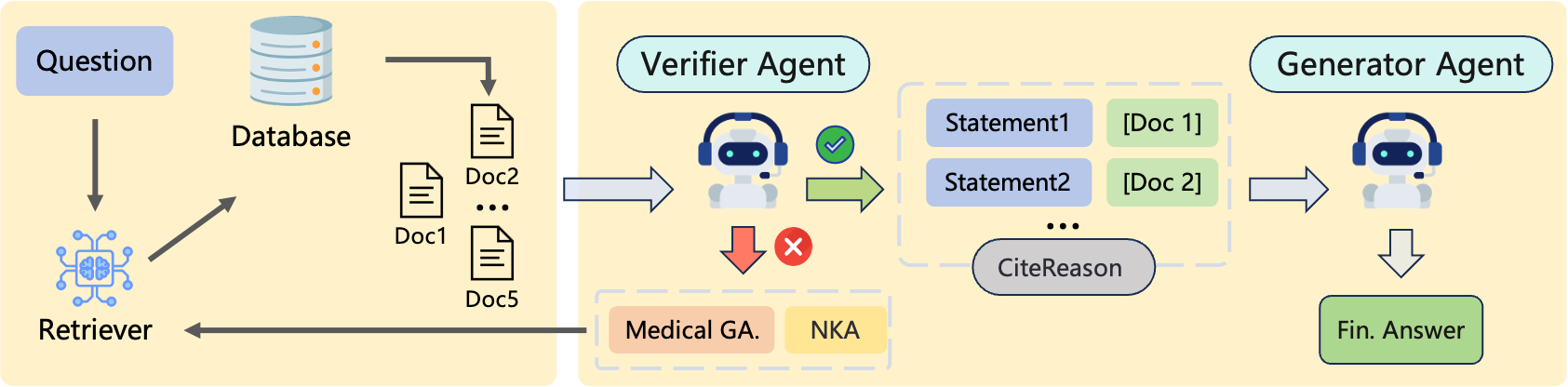}
\caption{
Illustration of the iterative retrieval-verification pipeline with dual-agent coordination. The verifier agent evaluates citation-grounded reasoning and provides Negative Knowledge Assertions (NKA) and Medical Gap Analysis (GA) for query refinement. The generator produces clinically grounded responses only from verified reasoning.
}
\label{f-2}
\end{figure*}

\begin{figure*}[t]
\centering
\includegraphics[width=1\textwidth]{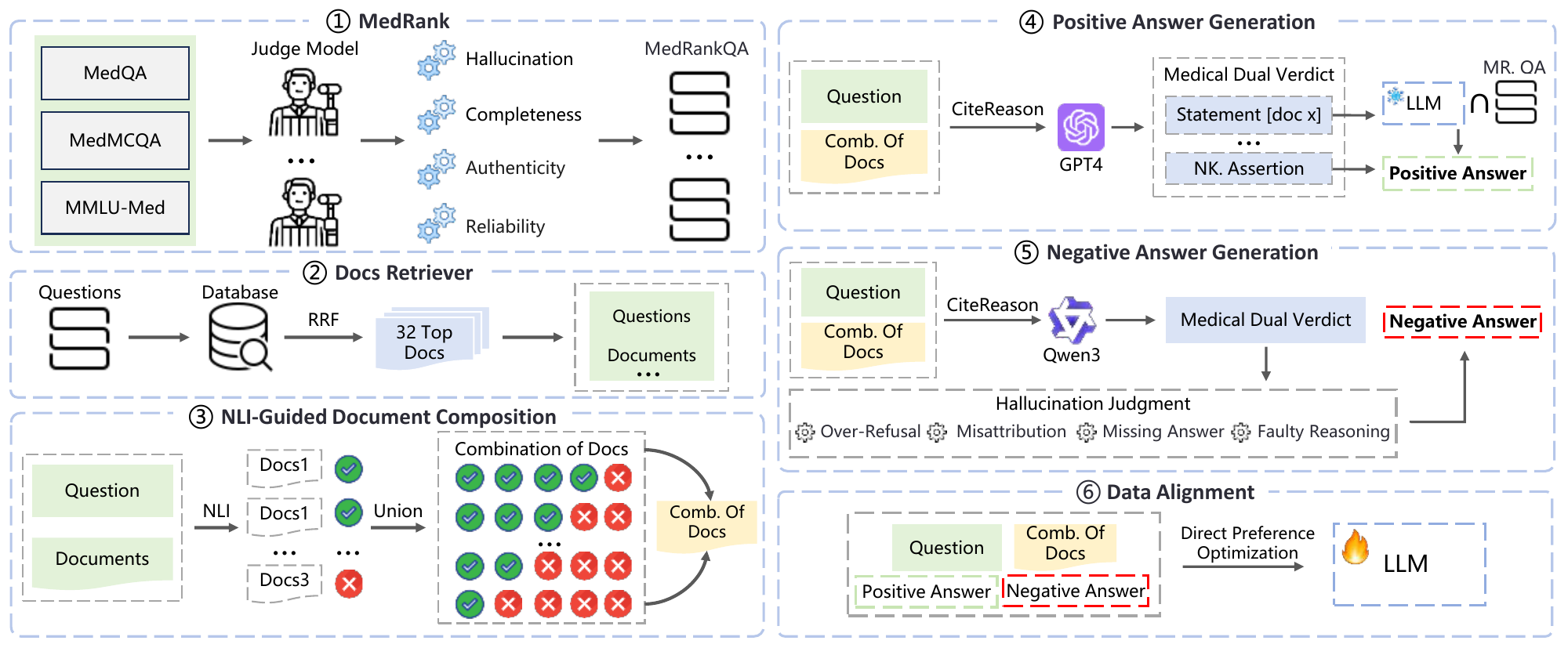}
\caption{
Overview of the MedTrust-Align framework for hallucination-aware medical question answering. The left pipeline depicts the knowledge construction process, including difficulty-aware sampling from MedRankQA (MR. QA), hybrid document retrieval, and NLI-based knowledge composition. The right pipeline shows the alignment training process, which involves positive sample generation from validated CiteReason and medically grounded NK. Assertion (Negative Knowledge Assertion), hallucination-aware negative sampling based on reasoning conflicts, and preference optimization through DPO training.
}
\label{f-3}
\end{figure*}

We start by defining the task of citation-aware reasoning for biomedical contexts. Next, we provide a detailed breakdown of our framework, focusing on the \textbf{Iterative Retrieval and Verification Pipeline}, where two agents improve evidence sufficiency. Additionally, we introduce the \textbf{MedTrust-Align Module (MTAM)}, a training framework that preserves medically grounded reasoning through evidence gap detection and knowledge supplementation. Fig.~\ref{f-2} presents an overview of the entire pipeline, and Fig.~\ref{f-3} illustrates the MTAM framework.

\subsection{Task Formulation}
\label{sec:task}

Given a medical query $q$, candidate answers $\mathcal{A} = \{a_1, a_2, \ldots, a_n\}$, and retrieved documents $\mathcal{D} = \{d_1, d_2, \ldots, d_m\}$, our objective is to identify the optimal answer from $\mathcal{A}$ through  citation-grounded reasoning statements (CiteReason). The model synthesizes a comprehensive statement $\mathcal{S} = \{s_1, s_2, \ldots\}$ where each statement $s_i$ is substantiated by empirical evidence and accompanied by precise inline citations $\mathcal{C}_i = \{c_{i,1}, c_{i,2}, \ldots\}$. Each citation $c_{i,j}$ corresponds to a specific source document $d_j \in \mathcal{D}$, represented as [Doc~$j$].

When the document collection $\mathcal{D}$ provides sufficient supporting evidence for generating a medically reliable response, the model proceeds to construct a citation-aware rationale in the format ``\textit{$statement_1$} [Doc~1] \textit{$statement_2$} [Doc~2]\textit{~$\ldots$}'', ensuring that each citation clearly corresponds to its respective source of evidence. In cases where such evidence is lacking, the model adheres to a principled refusal strategy by issuing a structured Negative Knowledge Assertion stating that \textit{``Insufficient evidence was identified in the retrieved content to support a medically reliable answer.''}

\subsection{Iterative Retrieval and Verification Pipeline}
\label{sec:pipeline}

We design an iterative retrieval and verification pipeline to address the limitations of static retrieval in biomedical question answering. This pipeline refines evidence using CiteReason and ensures that the retrieved content is both clinically accurate and complete. An overview is shown in Fig.~\ref{f-2}.

Given a query \(q\), the system retrieves an initial document set \(D = \{d_1, \ldots, d_n\}\) from a high-quality medical corpus \(\mathcal{K}\). The corpus includes PubMed abstracts\footnote{\url{https://pubmed.ncbi.nlm.nih.gov/}}, StatPearls\footnote{\url{https://www.statpearls.com/}}, standard medical Textbooks~\cite{jin2021disease}, and curated medical Wikipedia~\cite{xiong2024benchmarking}. A hybrid retrieval strategy combines sparse and dense methods to cover both lexical and semantic relevance:
\begin{equation}
D = \text{RRF}\left( \text{BM25}(q) \cup \text{Dense}_{\theta}(q) \right)
\label{eq:retrieval}
\end{equation}
Here, \(\text{BM25}(q)\) represents traditional keyword-based retrieval~\cite{robertson2009probabilistic}. \(\text{Dense}_{\theta}(q)\) refers to dense retrievers including MedCPT~\cite{jin2023medcpt} and Contriever~\cite{izacard2021unsupervised}. The results are merged and ranked using Reciprocal Rank Fusion (RRF)~\cite{cormack2009reciprocal}. The top 32 documents form the initial evidence set \(\mathcal{D}^{(0)}\).

A verifier agent \(\phi\), based on the MedTrust-Align Module, evaluates whether the document set \(\mathcal{D}^{(t)}\) supports valid citation-grounded reasoning. If the evidence is incomplete or unsupported, the verifier generates a structured Medical Gap Analysis \(\mathcal{M}^{(t)}\) along with Negative Knowledge Assertions. These guide query refinement:
\begin{equation}
q^{(t+1)} = \text{Augment}(q, \mathcal{M}^{(t)})
\label{eq:query_aug}
\end{equation}
An updated evidence set \(\mathcal{D}^{(t+1)}\) is then retrieved as \(R(q^{(t+1)}, \mathcal{K})\). This iterative process continues until the verifier produces a valid reasoning chain \(\mathcal{S}_{\text{valid}}\) or the iteration reaches the maximum step \(T_{\text{max}} = 3\).

After successful verification, a generator agent \(\psi\) produces a clinically grounded answer based on \(\mathcal{S}_{\text{valid}}\):
\begin{equation}
a = \psi(\mathcal{S}_{\text{valid}})
\label{eq:response_generation}
\end{equation}
If verification fails in all iterations, the system falls back on internal parametric reasoning to generate the response.

\subsubsection{Dual-Agent Coordination}

The pipeline relies on coordination between a verifier agent and a generator agent. The verifier \(\phi\) inspects the logical and clinical validity of retrieved content. If the evidence is not sufficient, it outputs Negative Knowledge Assertions and \(\mathcal{M}^{(t)}\) to guide the next query. If the evidence is verified, it returns CiteReason.

The generator \(\psi\) is activated only when verification succeeds. It generates the final answer based on \(\mathcal{S}_{\text{valid}}\), ensuring that the response is accurate and grounded in evidence.

\subsubsection{Targeted Evidence Refinement}

Whenever the verifier identifies a missing or weak component, the system uses \(\mathcal{M}^{(t)}\) to refine the query through Eq.~\ref{eq:query_aug}. This allows the retriever to focus on the gaps rather than repeating full retrieval. The loop continues until a valid reasoning path is found or the iteration limit is reached. If the evidence remains insufficient, the generator responds using parametric reasoning guided by internal knowledge.

\subsection{MedTrust-Align Framework}
\label{sec:framework}

\subsubsection{MedRankQA Dataset Construction}

We construct the \textbf{MedRankQA} dataset by merging the training sets of MedQA and MedMCQA, resulting in a combined corpus $\mathcal{D}_t$ containing over 180{,}000 QA pairs. Each query $q_i \in \mathcal{D}_t$ undergoes $k$ rounds of hallucination-aware self-assessment using varied decoding parameters (temperature, top-$k$, top-$p$) and a dedicated self-evaluation prompt $P_e$. In each round, the model is assessed against $k$ evaluation criteria:

\begin{equation}
\mathcal{E} = \{e_1, e_2, \ldots, e_k\}
\label{eq:eval_criteria}
\end{equation}

\noindent
where each $e_j$ targets a distinct reliability dimension, including hallucination, authenticity, completeness, and overall reliability. The final difficulty level $l_i$ for question $q_i$ is determined by the consistency of responses across all rounds. Based on $l_i$, the dataset is divided into three disjoint subsets.

The Stable Group $Q_s$ includes questions that are answered correctly in all $k$ rounds, demonstrating complete agreement with the ground truth. These represent the lowest difficulty level ($l_i = 0$) and reflect reliable model behavior under varied decoding conditions.

The Challenging Group $Q_h$ contains questions that are answered incorrectly in every round, showing no agreement with the ground truth. These correspond to the highest difficulty level ($l_i = k$) and are typically associated with ambiguous content or hallucination-prone inputs.

The Medium Group $Q_m$ consists of questions with partial agreement, where correct answers appear in some but not all rounds ($0 < l_i < k$). This group represents intermediate difficulty and highlights variability in model reliability.

\subsubsection{NLI-Guided Document Composition}

To construct reliable and contextually aligned training instances, we guide document selection using a natural language inference (NLI) model~\cite{poliak2018hypothesis}. This process operates on the candidate document set \(D = \{d_1, \ldots, d_n\}\), retrieved from \(\mathcal{K}\) via the hybrid retrieval strategy described above.

We adopt the T5-XXL-True-NLI-Mixture~\footnote{\url{https://huggingface.co/google/t5_xxl_true_nli_mixture}}  to assess whether each document \(d_j\) provides sufficient support for inferring the answer \(a\) given the question \(q\). For each \(d_j \in D\), we compute:

\begin{equation}
\text{NLI}(d_j, q, a) \rightarrow y_j \in \{\text{entail}, \text{not\_entail}\}
\label{eq:nli}
\end{equation}

Here, a label of \textit{entail} indicates that the document supports the inference from question to answer, while \textit{not\_entail} denotes insufficient or irrelevant evidence.

We then construct multiple five-document subsets with diverse entailment compositions to augment the training set, reflecting realistic retrieval scenarios with mixed-quality evidence.

\subsubsection{Positive Sample Construction}

Our positive sample construction methodology leverages self-assessed difficulty annotations from MedRankQA to generate trustworthy training datas. Each instance follows the Medical Dual Verdict formulation:

\begin{equation}
\mathcal{V} = \mathcal{R} \cup \mathcal{N}
\label{eq:dual_verdict}
\end{equation}

\noindent
where $\mathcal{R}$ denotes validated CiteReason and $\mathcal{N}$ represents Negative Knowledge Assertions.

Focusing on challenging queries $Q_h$ consistently associated with hallucinations, we apply GPT-4~\cite{achiam2023gpt} to synthesize initial reasoning over retrieved documents $D$, forming preliminary $\mathcal{R}$ and $\mathcal{N}$ sets. To validate reasoning quality, we implement a controlled verification process using a frozen Biomedical Response Generator \(\psi\). Each reasoning statement $r_i \in \mathcal{R}$ is processed as:

\begin{equation}
\psi(q, r_i) \rightarrow a_i
\label{eq:verify_positive}
\end{equation}

\noindent
where $a_i$ is the model-generated answer. If $a_i$ is correct, the tuple $(q, D, r_i)$ is retained as a verified positive sample.

To enhance diversity, we extend this verification across queries of all difficulty levels $Q_s$, $Q_m$, and $Q_h$. Moreover, legitimate Negative Knowledge Assertions $n_i \in \mathcal{N}$, explicitly generated by GPT-4 in response to evidence insufficiency, are preserved as positive samples capturing valid refusal behavior.

\subsubsection{Hallucination-Aware Negative Sample Construction}

We construct negative training instances that target four representative hallucination categories in biomedical question answering. To capture realistic hallucination patterns, our approach leverages multi-model collaboration to generate diverse negative samples.

The Faulty Reasoning category ($\mathcal{H}_F$) includes medical reasoning chains that exhibit logical inconsistencies or unsupported inferential leaps. For each query-document pair $(q, D)$ initially processed by GPT-4, we prompt Qwen3-4B to generate alternative CiteReason statements $r'$. We then apply NLI validation using T5-XXL-True-NLI-Mixture to verify whether $r'$ is entailed by $D$. If no entailment is detected, we construct the negative sample as:
\begin{equation}
\text{NLI}(r', D) = 0 \Rightarrow (q, D, r') \in \mathcal{H}_F
\label{eq:faulty}
\end{equation}
\noindent
where $\text{NLI}(r', D)$ indicates whether $r'$ is supported by $D$ under natural language inference.

The Missing Answer category ($\mathcal{H}_M$) focuses on questions from the easy subset $Q_s$, where \(\psi\) consistently answers correctly using parametric knowledge alone. CiteReason statements $r'$ are generated by Qwen3-4B following the same prompting strategy as GPT-4. If the model fails to answer $q$ correctly when only provided with $r'$, the sample is labeled as:
\begin{equation}
\psi(q \mid r') \neq \psi(q) \Rightarrow (q, D, r') \in \mathcal{H}_M
\label{eq:missing}
\end{equation}
\noindent
where $\psi(q \mid r')$ is the model's answer to $q$ when conditioned on $r'$, and $\psi(q)$ is its original response without additional reasoning input.

The Over-Refusal category ($\mathcal{H}_O$) includes cases where GPT-4 generates a valid rationale $r$ from $D$, while Qwen3-4B produces a Negative Knowledge Assertion $n_i$ despite the presence of sufficient evidence. After confirming the validity of $r$ through NLI and verifying that $\psi$ produces consistent answers with and without $r$, we construct the sample as:
\begin{align}
\text{NLI}(r, D) = 1 \;\land\; \psi(q \mid r) = 
\psi(q) \nonumber \\
\Rightarrow (q, D, n_i) \in \mathcal{H}_O
\label{eq:overrefusal}
\end{align}
\noindent
where $n_i$ is the refusal statement from Qwen3-4B, and $r$ is a valid rationale generated by GPT-4 that aligns with $D$ and supports the correct answer.

The Misattribution category ($\mathcal{H}_A$) addresses situations where CiteReason statements $r$ are paired with semantically similar but factually misaligned document sets $D'$. Using MedCPT-Article-Encoder~\footnote{\url{https://huggingface.co/ncbi/MedCPT-Article-Encoder}}, we retrieve distractor documents $D'$ that exhibit high semantic similarity to $D$ but do not entail $r$. A misattribution sample is created when:
\begin{equation}
\text{Sim}(D, D') > \delta \land \text{NLI}(r, D') = 0 \Rightarrow (q, D', r) \in \mathcal{H}_A
\label{eq:misattribute}
\end{equation}
\noindent
where $\text{Sim}(D, D')$ measures the semantic similarity between $D$ and $D'$, and $\delta$ is a predefined similarity threshold.

To capture hallucination behaviors across different levels of question complexity, we extend the construction of negative samples beyond the easy-question subset $Q_s$ to include medium- and hard-difficulty queries from $Q_m$ and $Q_h$. This difficulty-aware sampling strategy enhances the alignment module's ability to detect hallucinations across diverse clinical scenarios.

\subsubsection{Training Configuration and Optimization}

We construct a comprehensive training corpus $\mathcal{T}$ comprising over 17,000 meticulously annotated instances, strategically balanced between verified positive samples and hallucination-aware negative samples. The training corpus $\mathcal{T}$ consists of positive pairs $\mathcal{V}^+ = \{\mathcal{R}^+, \mathcal{N}^+\}$ and hallucination-based negatives $\mathcal{V}^- \subset \mathcal{H}$. The MedTrust-Align module $\phi$ is optimized using DPO:

\begin{align}
\mathcal{L}_{\text{DPO}} = 
- \mathbb{E}_{(q, D, \mathcal{V}^+, \mathcal{V}^-) \sim \mathcal{T}} \Big[
\log \sigma\big( 
\beta \log \tfrac{\pi_\theta(\mathcal{V}^+|q,D)}{\pi_{\text{ref}}(\mathcal{V}^+|q,D)} 
\notag \\ 
- \beta \log \tfrac{\pi_\theta(\mathcal{V}^-|q,D)}{\pi_{\text{ref}}(\mathcal{V}^-|q,D)} 
\big) \Big]
\label{eq:dpo_loss}
\end{align}

\noindent
where $\pi_\theta$ is the model policy, $\pi_{\text{ref}}$ is a reference model, $\beta$ controls preference sharpness, and $\sigma$ is the sigmoid function.

\section{EXPERIMENTS}
\subsection{Experimental Settings}
\subsubsection{Datasets}
We evaluate our framework using three widely adopted biomedical multiple-choice question answering datasets, namely MedMCQA, MedQA, and MMLU-Med. These benchmarks provide comprehensive assessment of both factual knowledge recall and clinical reasoning capabilities within open-domain medical contexts. For MMLU-Med, we focus on six medical and life science subdomains including anatomy, clinical knowledge, college biology, college medicine, medical genetics, and professional medicine.

\begin{table*}[htbp]
\small
\centering
\caption{Performance comparison on open-domain biomedical QA benchmarks}
\label{tab:main_results}
\setlength{\tabcolsep}{4.5pt}
\begin{tabular*}{\textwidth}{@{\extracolsep{\fill}}llcccc}
\toprule
Model & Method & MedMCQA (\%) & MedQA (\%) & MMLU-Med (\%) & Average (\%) \\
\midrule
Self-BioRAG$^\dag$         & -              & 44.0          & 48.6          & 57.2          & 49.9 \\
Med-PaLM$^\dag$            & -              & 56.5          & 60.3          & 75.6          & 64.1 \\
GPT-3.5$^\dag$             & -              & 51.0          & 53.6          & 67.3          & 57.3 \\
GPT-4-base$^\dag$          & -              & \textbf{73.7}          & \textbf{86.1}          & \textbf{89.9}          & \textbf{83.2} \\
\midrule
\multirow{9}{*}{LLaMA3.1-8B-Instruct}
                  & Zero-Shot    & 47.7          & 51.9          & 63.5          & 54.4 \\
                  & COT          & 52.9          & 59.3          & 67.4          & 59.9 \\
                  & RAG          & 53.3          & 59.6          & 70.2          & 61.0 \\
                  & GenRead      & 53.8          & 59.0          & 71.6          & 61.4 \\
                  & PostAttr     & 53.7          & 61.6          & 68.8          & 61.3 \\
                  & ICL          & 54.5          & 58.4          & 68.4          & 60.4 \\
                  & Summary      & 50.5          & 57.4          & 67.3          & 58.4 \\
                  & Ours (SFT)   & 54.7          & \textbf{63.1} & 69.0          & 62.3 \\
                  & Ours (DPO)   & \textbf{57.5} & 62.3          & \textbf{72.4} & \textbf{64.1} \\
\midrule
\multirow{9}{*}{Qwen3-8B}
                  & Zero-Shot    & 58.0          & 62.6          & 78.9          & 66.5 \\
                  & COT          & 60.8          & 67.5          & 79.8          & 69.4 \\
                  & RAG          & 59.7          & 68.9          & 81.3          & 69.9 \\
                  & GenRead      & 60.0          & 68.7          & 81.9          & 70.2 \\
                  & PostAttr     & 60.3          & 67.6          & 82.0          & 69.9 \\
                  & ICL          & 61.3          & 66.4          & 81.1          & 69.6 \\
                  & Summary      & 61.4          & 64.5          & 80.9          & 68.9 \\
                  & Ours (SFT)   & 62.8          & 68.7          & 82.5          & 71.3 \\
                  & Ours (DPO)   & \textbf{63.6} & \textbf{70.1} & \textbf{84.3} & \textbf{72.6} \\
\bottomrule
\end{tabular*}
\vspace{2mm}

\footnotesize\raggedright
{Notes: Results are reported on three biomedical QA benchmarks including MedMCQA, MedQA, and MMLU-Med. Exact Match is used to measure answer accuracy and is expressed in percentage (\%). $^\dag$Results are adopted from Jeong et al.~\cite{jeong2024improving}. Bold numbers indicate the best performance on each benchmark.}
\end{table*}

\subsubsection{Baselines}
We compare our approach against several competitive baseline methods. These include Self-BioRAG~\cite{jeong2024improving}, which implements the original RAG pipeline using LLaMA2-13B~\cite{touvron2023llama}, Med-PaLM~\cite{singhal2023towards}, a domain-specific language model developed by Google, GPT-3.5~\cite{openai2023a_chatgpt} and GPT-4-base~\cite{achiam2023gpt}, as well as two recent instruction-tuned open-source models, LLaMA3.1-8B-Instruct~\cite{touvron2024llama3} and Qwen3-8B~\cite{qwen32024}.

The open-source models are evaluated across six distinct QA paradigms. In-Context Learning (ICL)~\cite{gao2023enabling} incorporates three demonstrations for each query, with each demonstration comprising a sample question, its corresponding top-5 retrieved documents, and inline cited statements. Chain-of-Thought (CoT)~\cite{wei2022chain} utilizes CoT prompting to enable step-by-step reasoning. RAG~\cite{xiong2024benchmarking} represents a standard retrieval-augmented generation framework. GenRead~\cite{yu2022generate} follows a generate-then-retrieve approach that first hypothesizes an answer, retrieves supporting evidence, and subsequently generates the final answer. PostAttr~\cite{gao2023retrieval} performs post-hoc citation attribution by first generating an answer and then employing an NLI model to select supporting evidence from the top-5 retrieved documents. Summary~\cite{vig2021exploring} generates answers conditioned on summaries of retrieved documents.

All retrieval-based baselines, including RAG, GenRead, PostAttr, ICL and Summary, utilize the same retrieval corpus and methodology employed in our proposed approach. Specifically, they retrieve from the MedRankQA corpus described in Section~\ref{sec:pipeline} using an identical retrieval pipeline to ensure fair comparison.

\subsubsection{Evaluation Metrics}
We report exact match (EM) as the primary evaluation metric~\cite{rajpurkar2016squad}. An answer is considered correct only if it exactly matches the ground-truth option. This metric is widely adopted in biomedical QA tasks and serves as a reliable measure of accuracy in multiple-choice settings.

\subsection{Main Results}
\label{sec:main_results}

Table~\ref{tab:main_results} presents the comprehensive evaluation of our MedTrust-Align framework on three open-domain biomedical QA benchmarks. Our approach demonstrates significant advantages over existing methods through two key innovations, namely the iterative retrieval-verification mechanism and the MedTrust-Align strategy. The MedTrust-Align (DPO) variant achieves substantial improvements of 64.1~\% average EM on LLaMA3.1-8B-Instruct, representing 9.7 percentage points improvement over zero-shot and 2.7 percentage points over the best RAG baseline. On Qwen3-8B, our method achieves 72.6~\% average EM with 6.1 percentage points improvement over zero-shot and 2.4 percentage points over the best RAG baseline. These gains fundamentally stem from our iterative retrieval mechanism that transforms static, one-shot retrieval into dynamic, feedback-driven evidence construction. Traditional RAG methods rely solely on initial retrieval quality, often suffering from insufficient or noisy evidence that leads to unreliable medical reasoning. In contrast, our dual-agent system employs the Verifier Agent to systematically identify evidence insufficiencies and generate Medical Gap Analysis. This analysis enables the system to iteratively refine retrieval queries and progressively fill knowledge gaps until valid citation-grounded reasoning is achieved. Our MedTrust-Align training strategy systematically teaches models to distinguish between reliable medical reasoning and various forms of hallucinated responses through carefully constructed training instances that span diverse clinical scenarios and question complexities. The consistent improvements observed when transitioning from SFT to DPO across all benchmarks demonstrate the strength of our medical trust alignment approach. For instance, on MMLU-Med with LLaMA3.1, accuracy improves from 69.0~\% to 72.4~\%. This shows that our alignment strategy helps models produce medically sound reasoning and reduces unreliable answers. The preference optimization enabled by DPO allows the model to learn from both reliable, citation-supported reasoning and negative hallucination examples.

\begin{figure*}[!t]
\centering
\hspace*{-3em}\includegraphics[width=2.25\columnwidth]{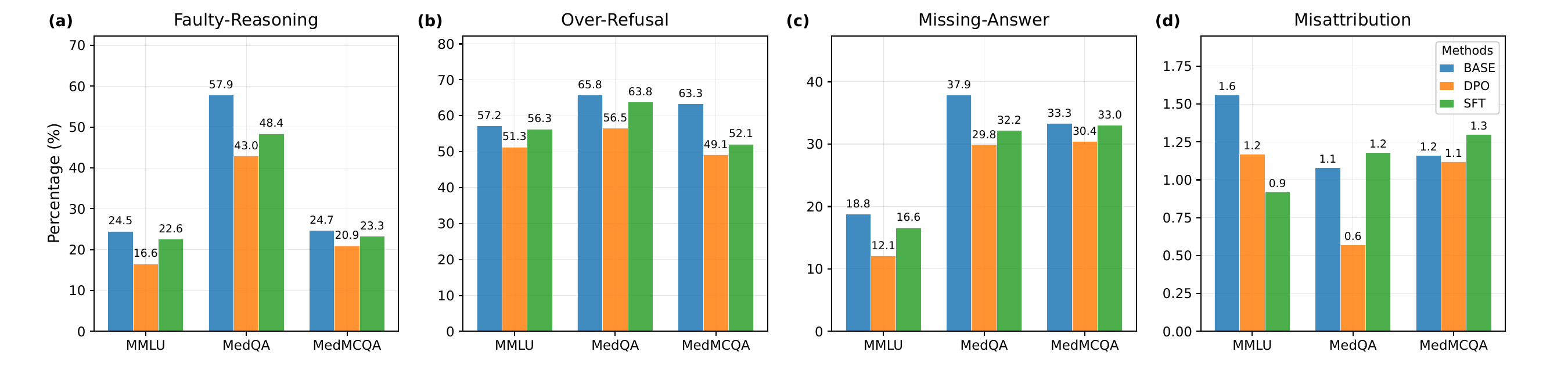}
\caption{
Distribution of four hallucination types across the MMLU, MedQA and MedMCQA datasets. Subfigures (a)–(d) show Faulty Reasoning, Over Refusal, Missing Answer and Misattribution, respectively. Each bar indicates the proportion under three model variants, where BASE is the original model without alignment, DPO is trained with preference optimization and SFT is trained with supervised fine-tuning.
}

\label{f-4}
\end{figure*}

\subsection{Ablation Study}
\label{sec:ablation}

We perform ablation experiments to evaluate the contribution of each component in the MedTrust-Guided Iterative RAG framework. Table~\ref{tab:ablation} summarizes the performance changes when removing key modules.

\textbf{Effect of MTAM.} Removing the Medical Trust-Align Module (w/o MTAM) results in significant performance degradation across all benchmarks. The system reverts to the base language model for generating reasoning statements and Negative Knowledge Assertions without DPO-based refinement. LLaMA3.1-8B-Instruct shows decreases of 2.3\%, 3.8\%, and 4.3\% on MedMCQA, MedQA, and MMLU-Med. Qwen3-8B records corresponding drops of 1.2\%, 3.2\%, and 3.6\%. These results confirm that MTAM enhances factual accuracy by guiding the model to produce CiteReason and properly handle evidence insufficiency. Without MTAM, the model lacks hallucination-aware supervision and struggles to differentiate between supported and unsupported claims.

\begin{table}[!t]
\caption{Ablation Study on Biomedical QA Benchmarks}
\label{tab:ablation}
\setlength{\tabcolsep}{3.5pt}
\centering
\begin{tabular}{@{}lp{2.2cm}ccc@{}}
\toprule
\textbf{Model} & \textbf{Method} & \textbf{MedMCQA} & \textbf{MedQA} & \textbf{MMLU-Med} \\
\midrule
\multirow{3}{*}{LLaMA3.1-8B} 
  & Ours (DPO)      & \textbf{57.5} & \textbf{62.3} & \textbf{72.4} \\
  & w/o MTAM        & 55.2 & 58.5 & 68.1 \\
  & w/o IR          & 56.1 & 60.8 & 69.0 \\
  &w/o MTAM and IR  & 54.1 & 57.2 & 67.7          \\
\midrule
\multirow{3}{*}{Qwen3-8B} 
  & Ours (DPO)      & \textbf{63.6} & \textbf{70.1} & \textbf{84.3} \\
  & w/o MTAM        & 62.4 & 66.9 & 80.7 \\
  & w/o IR          & 62.5 & 68.8 & 82.5 \\
  & w/o MTAM and IR & 61.7 & 65.8 & 78.5   \\
\bottomrule
\end{tabular}
\vspace{1mm}

\noindent
\begin{minipage}{\linewidth}
\footnotesize\raggedright
{Note: Ablation experiments are conducted by removing key components from the MedTrust-Align framework: the MedTrust-Align Module (MTAM) and the Iterative Retrieval mechanism (IR). Results are reported using the Exact Match metric (\%). Bold values indicate the best result.}
\end{minipage}
\end{table}

\textbf{Impact of Iterative Retrieval.} Excluding the iterative retrieval module (w/o IR) leads to performance declines between 1.1\% and 3.4\%. The framework then relies solely on MTAM's initial judgment without further evidence refinement. The largest decrease is observed on MMLU-Med with a drop of 3. 4\% for LLaMA3.1-8B-Instruct, suggesting that iterative retrieval is especially beneficial for complex queries. Iterative retrieval improves the completeness of evidence and supports deeper reasoning.

\textbf{Combined Effects.} Removing both MTAM and IR (w/o MTAM and IR) yields the most substantial degradation. LLaMA3.1-8B-Instruct drops by 3.4\%, 5.1\%, and 4.7\% on MedMCQA, MedQA, and MMLU-Med. This configuration lacks both evidence refinement and guided reasoning. Although MTAM provides the larger gain, the two modules contribute complementary strengths. MTAM improves reasoning accuracy while iterative retrieval enhances evidence sufficiency.

These findings indicate that both MTAM and IR are essential for reliable biomedical question answering. Each addresses a distinct challenge, and their combination enables more accurate and trustworthy model behavior.

\subsection{Hallucination Pattern Analysis}

To evaluate the factual consistency of model-generated reasoning, we analyze four hallucination types defined in Section~\ref{sec:framework}, namely Faulty Reasoning, Over-Refusal, Missing Answer, and Misattribution. We utilize the T5-XXL-True-NLI-Mixture model to assess entailment relationships among retrieved documents, generated statements, and question-answer pairs, thereby enabling scalable evaluation of factual alignment.

For each hallucination type, the measurement is defined as follows. Faulty Reasoning refers to generated statements that cannot be logically inferred from the cited documents based on natural language inference. Over-Refusal describes cases where the MTAM declines to produce an answer despite the availability of sufficient supporting evidence. Missing Answer captures instances in which the generated response lacks essential information required to derive the correct answer when interpreted by the response agent. Misattribution corresponds to incorrect citations where the referenced document does not support the statement, although other retrieved documents do provide relevant information.

Fig.~\ref{f-4} presents the distribution of these hallucination types across three datasets. Experimental results indicate that DPO outperforms other methods across most hallucination types and datasets, with comparable performance to SFT on Misattribution in MMLU. On MedQA, the rate of Faulty Reasoning decreases notably from 57.9~\% to 43.0~\%, reflecting improved logical consistency in medical reasoning. Over-Refusal is also substantially reduced, with the most significant improvement observed on MedMCQA, where the rate drops from 63.3~\% to 49.1~\%. This suggests that the model achieves a better balance between cautiousness and informativeness. Missing Answer occurrences decrease by 6.7~\% on MMLU-Med and by 8.1~\% on MedQA, indicating more complete and informative responses. Misattribution remains consistently low under all configurations, with DPO achieving the best result on MedQA at 0.6~\%.

Overall, the DPO-trained model consistently outperforms both the Base and SFT variants across most hallucination types, highlighting the advantage of preference optimization in learning from both successful and flawed reasoning behaviors. Although SFT yields moderate improvements over the base model, its gains are limited in the absence of explicit feedback on generation quality. The strong and robust performance of DPO across multiple datasets and reasoning conditions validates the effectiveness of the MedTrust-Align framework in reducing diverse hallucination patterns and enhancing factual reliability in biomedical question answering.

\section{Conclusion}

In this work, we introduced MedTrust-Guided Iterative RAG, a trust-aware retrieval-augmented generation framework designed to improve the reliability of biomedical question answering. By incorporating citation-grounded reasoning, iterative retrieval-verification, and hallucination-aware preference optimization, our method enhances the factual consistency and trustworthiness of generated responses. Through the MedTrust-Align Module and dual-agent coordination, the framework offers a robust solution for complex medical queries. Experimental evaluations on MedMCQA, MedQA, and MMLU-Med demonstrate substantial improvements, with average accuracy gains of 2.7\% and 2.4\% over the strongest standard RAG baselines. This work represents a meaningful step toward building safer and more interpretable AI systems for clinical decision support.

\section*{\small ACKNOWLEDGMENT}
This research was supported by the Natural Science Foundation of China (No. 62276043, 62302076), 
the Fundamental Research Funds for the Central Universities (No. DUT25YG108), and the Research Project on High Quality Development of Hospital Pharmacy, National Institute of Hospital Administration, NHC, China (No. NIHAYSZX2525).




\bibliographystyle{IEEEtran}
\bibliography{references} 

\end{document}